\documentclass{article}

\usepackage{microtype}
\usepackage{graphicx}
\usepackage{subcaption}
\usepackage{booktabs} 

\usepackage{hyperref}

\usepackage[preprint]{icml2026}

\usepackage{amsmath}
\usepackage{amssymb}
\usepackage{mathtools}
\usepackage{amsthm}

\usepackage[capitalize,noabbrev]{cleveref}

\usepackage{graphicx}
\usepackage{wrapfig}
\usepackage{cutwin}
\usepackage{subcaption}         
\usepackage{tabularx} 
\usepackage{array}
\usepackage{booktabs, multirow, caption} 
\usepackage[utf8]{inputenc} 
\usepackage[T1]{fontenc}    
\usepackage{hyperref}       
\usepackage{url}            
\usepackage{amsfonts}       
\usepackage{nicefrac}       
\usepackage{microtype}      
\usepackage[table]{xcolor}  
\usepackage{enumitem}
\usepackage{parskip}
\usepackage{tcolorbox}
\tcbuselibrary{listings}
\usepackage{gensymb}
\definecolor{myblue}{RGB}{216, 233, 246}

\theoremstyle{plain}

\theoremstyle{definition}

\theoremstyle{remark}

\usepackage[textsize=tiny]{todonotes}

\newcommand{\icmlfootnote}{\textsuperscript{*}Work done during the visiting at Adelaide University.}

\icmltitlerunning{SpatialAnt: Autonomous Zero-Shot Robot Navigation via Active Scene Reconstruction and Visual Anticipation}

\begin{document}

\twocolumn[
  \icmltitle{SpatialAnt: Autonomous Zero-Shot Robot Navigation via \\
  Active Scene Reconstruction and Visual Anticipation}
  


  \icmlsetsymbol{footnote}{*}

  \begin{icmlauthorlist}
    \icmlauthor{Jiwen Zhang}{fdu,aiml,footnote}
    \icmlauthor{Xiangyu Shi}{aiml}
    \icmlauthor{Siyuan Wang}{usc}
    \icmlauthor{Zerui Li}{aiml}
    \icmlauthor{Zhongyu Wei}{fdu,sh}
    \icmlauthor{Qi Wu}{aiml}\\
    \href{https://imnearth.github.io/Spatial-X/}{\texttt{Homepage: https://imnearth.github.io/Spatial-X/}}
  \end{icmlauthorlist}

  \icmlaffiliation{fdu}{Fudan University}
  \icmlaffiliation{usc}{University of Southern California}
  \icmlaffiliation{aiml}{Adelaide University}
  \icmlaffiliation{sh}{Shanghai Innovation Institute}
  \icmlcorrespondingauthor{Zhongyu Wei}{zywei@fudan.edu.cn}


  \vskip 0.3in
]



\printAffiliationsAndNotice{\icmlfootnote}  

\begin{abstract}
Vision-and-Language Navigation (VLN) has recently benefited from Multimodal Large Language Models (MLLMs), enabling zero-shot navigation. While recent exploration-based zero-shot methods~\cite{zhang2026spatialnav} have shown promising results by leveraging global scene priors, they rely on high-quality human-crafted scene reconstructions, which are impractical for real-world robot deployment.
When encountering an unseen environment, a robot should build its own priors through pre-exploration. However, these self-built reconstructions are inevitably incomplete and noisy, which severely degrade methods that depend on high-quality scene reconstructions.
To address these issues, we propose \textbf{SpatialAnt}, a zero-shot navigation framework designed to bridge the gap between imperfect self-reconstructions and robust execution. SpatialAnt introduces a physical grounding strategy to recover the absolute metric scale for monocular-based reconstructions. Furthermore, rather than treating the noisy self-reconstructed scenes as absolute spatial references, we propose a novel visual anticipation mechanism. This mechanism leverages the noisy point clouds to render future observations, enabling the agent to perform counterfactual reasoning and prune paths that contradict human instructions. Extensive experiments in both simulated and real-world environments demonstrate that SpatialAnt significantly outperforms existing zero-shot methods. We achieve a 66\% Success Rate (SR) on R2R-CE and 50.8\% SR on RxR-CE benchmarks. Physical deployment on a Hello Robot further confirms the efficiency and efficacy of our framework, achieving a 52\% SR in challenging real-world settings.

\end{abstract}

\section{INTRODUCTION}
\label{sec:intro}

\begin{figure}[t]
\setlength{\abovecaptionskip}{-1pt}
\setlength{\belowcaptionskip}{0pt}
    \centering
    \includegraphics[width=\linewidth]{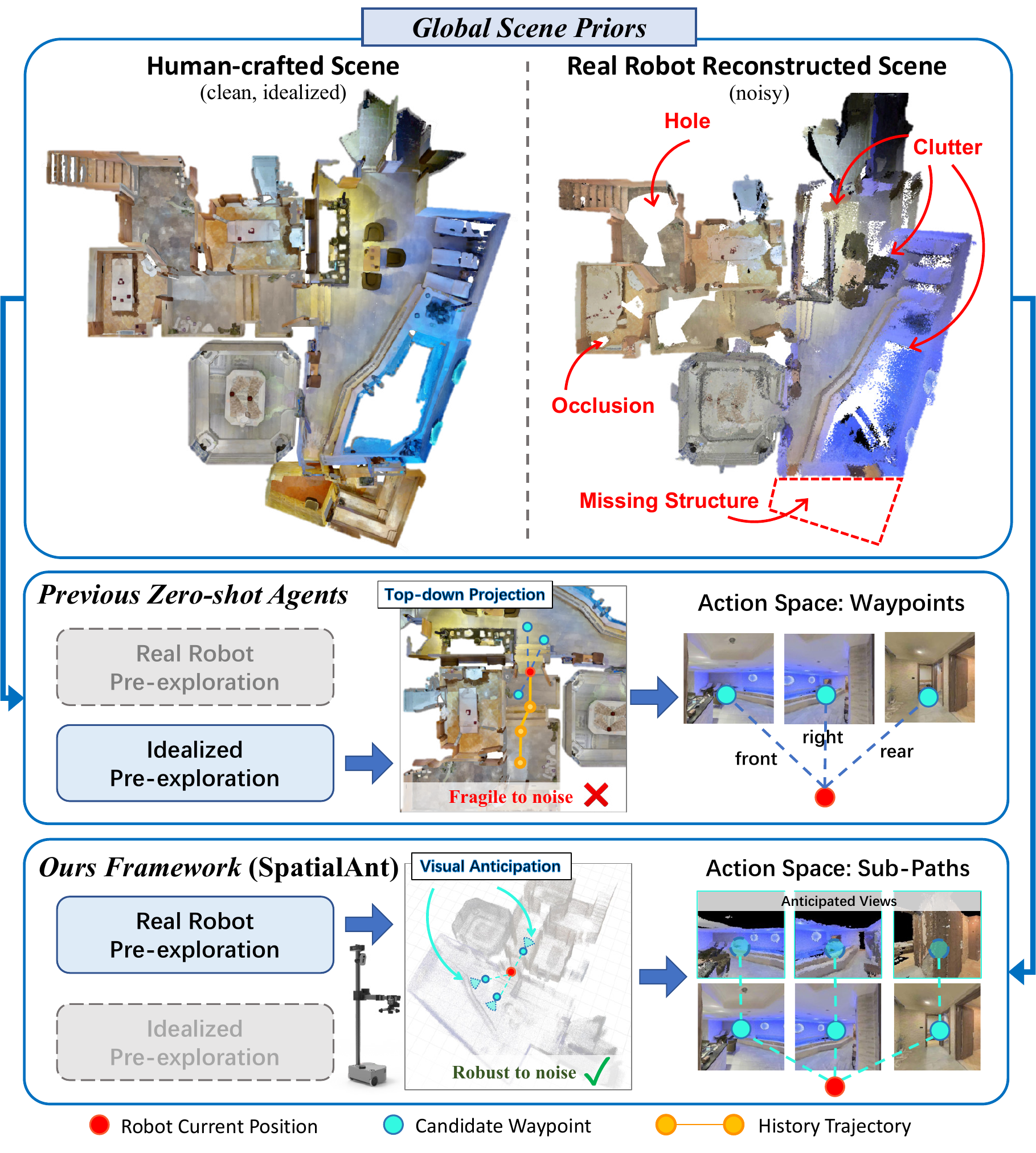}
    \caption{\textbf{The reality gap for pre-exploration based zero-shot VLN agents.}
    Previous works adopt the ideal scene reconstructions provided by existing datasets~\cite{chang2017matterport3d}, treating these global priors as accurate maps, and are sensitive to noise. Instead, we use the real robot reconstructed noisy scenes and propose a visual anticipation approach to transform the noisy scene into supportable sub-paths.
    }
    \label{fig:intro}
\end{figure}

Vision-and-Language Navigation (VLN)~\cite{anderson2018vision} is a fundamental challenge in embodied AI that requires agents to follow human instructions while navigating real environments~\cite{gu2022vision}. With the rapid development of Multimodal Large Language Models (MLLMs)~\cite{openai2025gpt5systemcard,comanici2025gemini}, which possess strong cross-modal reasoning ability and open-world knowledge, MLLM-based zero-shot VLN agents have emerged~\cite{qiao2025open,shi2025smartway,he2025strider}.
Recent advances in zero-shot VLN methods~\cite{bhatt2025vln,zhang2026spatialnav} have demonstrated the importance of pre-exploring the environment and obtaining global scene priors for navigation. These approaches, especially~\cite{zhang2026spatialnav}, make the idealized assumption that high-quality scene reconstructions, such as the perfect point cloud of the environment in Matterport3D dataset~\cite{chang2017matterport3d}, are readily available after pre-exploration. This introduces two significant reality gaps when a real robot is deployed in an unseen environment to actively explore the scene. 

Firstly, the acquisition of physically-grounded scene reconstructions is difficult under real robot deployment. Many robots are equipped only with monocular RGB cameras to maintain lightweight and low-cost. 
Even when depth sensors are present, they are often constrained by considerable noise and range limitations, compromising traditional depth-dependent methods such as TSDF~\cite{curless1996volumetric}. 
While recent neural SLAM approaches~\cite{wang2024dust3r, liu2025slam3r} have attempted 3D reconstruction solely from RGB sequences, the absence of absolute depth leads to a fundamental scale ambiguity of the reconstructed scene. Without a reliable metric scale, the self-reconstructed scene fails to serve as the navigation reference. 

Secondly, reconstructed scenes from the egocentric robotic observations are inevitably subjected to noise, as shown in Figure~\ref{fig:intro}. Unlike human-captured exocentric scene reconstructions in standard datasets~\cite{chang2017matterport3d}, egocentric perception of robots is constrained by their physical embodiment, such as lower camera mounting heights and restricted field of view, leading to incomplete spatial coverage and structural occlusions in the reconstructed point clouds. 
This discrepancy causes previous exploration-based zero-shot agents to degrade significantly when the noisy scene priors are used, posing a dilemma: \textit{if a self-reconstructed scene is destined to be noisy, can it still reliably guide a zero-shot agent?} 

To address these challenges, we present \textbf{SpatialAnt}, an autonomous zero-shot navigation framework that bridges the gap between imperfect reconstructions and robust execution. Specifically, SpatialAnt adopts a practical setting to actively explore the environment using a single RGB camera. 
To resolve the scale ambiguity of monocular vision, we incorporate \textbf{a physical grounding strategy} that anchors the self-reconstructed point cloud to the metric scale in real-world units (meters), through a comparison between the rendered depth from the point cloud and the predicted metric depth from Depth-Anything-2~\cite{yang2024depth}.
Since the reconstructed scenes are inherently noisy, in contrast to previous works~\cite{zhang2026spatialnav}, we propose not to treat them simply as an absolute position reference, but rather as a foundation for visual alignment. Analogous to biological ants utilizing their antennae to anticipate the path ahead, we propose \textbf{a novel visual anticipation mechanism}, that leverages the noisy point clouds to render the future view image for each action candidate. As shown in Figure~\ref{fig:intro}, the rendered expectations are combined with the agent's current observations to reformulate the action space as sub-paths, which constitute the embodied basis for counterfactual reasoning, allowing the agent to prune paths that visually contradict the instructions. 

Through extensive experiments in both simulated and real-world environments, we demonstrate that the self-constructed noisy scenes can still serve as a reliable navigation prior knowledge under the visual anticipation mechanism, where SpatialAnt establishes a new zero-shot state-of-the-art, achieving 66\% Success Rate (SR) in R2R-CE and 50.8\% SR in RxR-CE benchmarks, improving the previous best zero-shot baseline by +2\% and +18.4\% SR, respectively. 
Moreover, SpatialAnt reaches 52\% SR when deployed on the Hello Robot~\cite{HelloRobot2025}, demonstrating strong sim-to-real transfer ability for challenging real-world tasks. 

Our contributions are summarized below:
\begin{itemize}
\item We propose \textbf{SpatialAnt}, a zero-shot VLN framework that seamlessly integrates active exploration, physically-grounded scene reconstruction and anticipation-guided navigation, achieving state-of-the-art performance in both simulated and real environments.
\item We introduce a physical grounding strategy that recovers the physical metric for monocular RGB-based reconstructions, ensuring metric-consistent scene representations as the reliable navigation foundation.
\item We develop a novel visual anticipation mechanism that treats noisy scene reconstructions as verifiable visual cues, enabling zero-shot agents to perform counterfactual reasoning in anticipated physical space.
\end{itemize}

\section{RELATED WORK}
\label{sec:related_work}

\paragraph{\textbf{Vision-and-Language Navigation (VLN)}} requires an embodied agent to navigate by following natural language instructions~\cite{anderson2018vision,qi2020reverie}. Early studies
mainly focused on discrete environments~\cite{anderson2018vision,tan2019learning} where the action space is abstracted as a predefined topological graph. 
Recent trends have shifted towards continuous
environments (VLN-CE)~\cite{krantz2020beyond,hong2022bridging} to better align with real-world scenarios. Existing methods primarily adopt end-to-end learning paradigms on massive annotated datasets~\cite{hong2021vln,zhang2021curriculum,qiao2022hop}, and these approaches are further combined with MLLMs~\cite{zhou2024navgpt2,zhang2024navid,zhang2025embodied}. However, these learning-based methods are computationally expensive and generalize poorly to unseen environments.

\paragraph{\textbf{Zero-shot VLN}}
To address these limitations, zero-shot VLN agents are proposed~\cite{qiao2025open,shi2025smartway,he2025strider}, leveraging the reasoning capabilities of MLLMs to facilitate generalizable navigation. The fundamental challenge lies in providing sufficient embodied context for MLLMs to ground high-level semantic reasoning into actionable,  spatially-aware decisions~\cite{shi2025fast,li2026one}. 
To this end, Spatial-VLN~\cite{yue2026spatial} equips MLLM-based agents with explicit spatial attributes such as doors. 
However, the lack of persistent global structure awareness makes it struggle in long-horizon tasks. SpatialNav~\cite{zhang2026spatialnav} pushes this direction further by enabling pre-exploration and building a spatial scene graph to provide reusable global spatial priors. While effective, SpatialNav is primarily based on idealized point clouds provided by annotated datasets~\cite{chang2017matterport3d}. There is a substantial discrepancy between the noisy reconstructed scenes from the egocentric perception of real robots and these human-crafted surrogates. Motivated by this, we propose a novel visual anticipation mechanism to bridge the reality gap.

\paragraph{\textbf{Scene Representations in VLN}}
Current scene representations in VLN can be broadly categorized into on-the-fly and pre-explored structures. The prior typically construct topological maps~\cite{wang2021structured,chen2022think} or topometric memories~\cite{wang2023gridmm,zhang-etal-2025-mapnav} to expand the agent searchable action space. Recently, GTA~\cite{li2026one} uses an TSDF-reconstructed metric world built incrementally to decouple spatial modeling from semantic reasoning. However, these on-the-fly approaches suffer from the agent local line-of-sight, lacking necessary global context needed to resolve spatial ambiguities. In contrast, pre-exploration methods have emerged to establish a persistent global scene understanding before task execution. While VLN-Zero~\cite{bhatt2025vln} caches sparse local views and SpatialNav~\cite{zhang2026spatialnav} leverages spatial maps to underscore the importance of global geometry, a critical gap remains: none of these works have attempted full scene reconstruction after pre-exploration. 
This is largely due to the practical challenges of recoding precise pose and depth of the agent during exploration. 
To eliminate these constraints, we propose to leverage neural SLAM-based method to reconstruct dense 3D scenes from egocentric RGB streams. By incorporating a physical grounding strategy, to the best of our knowledge, we are the first to establish the closed-loop from active agent exploration to physically-grounded scene reconstruction.



\begin{figure*}
\vspace{0.1cm}
\setlength{\abovecaptionskip}{3pt}
\setlength{\belowcaptionskip}{0pt}
    \centering
    \includegraphics[width=\linewidth]{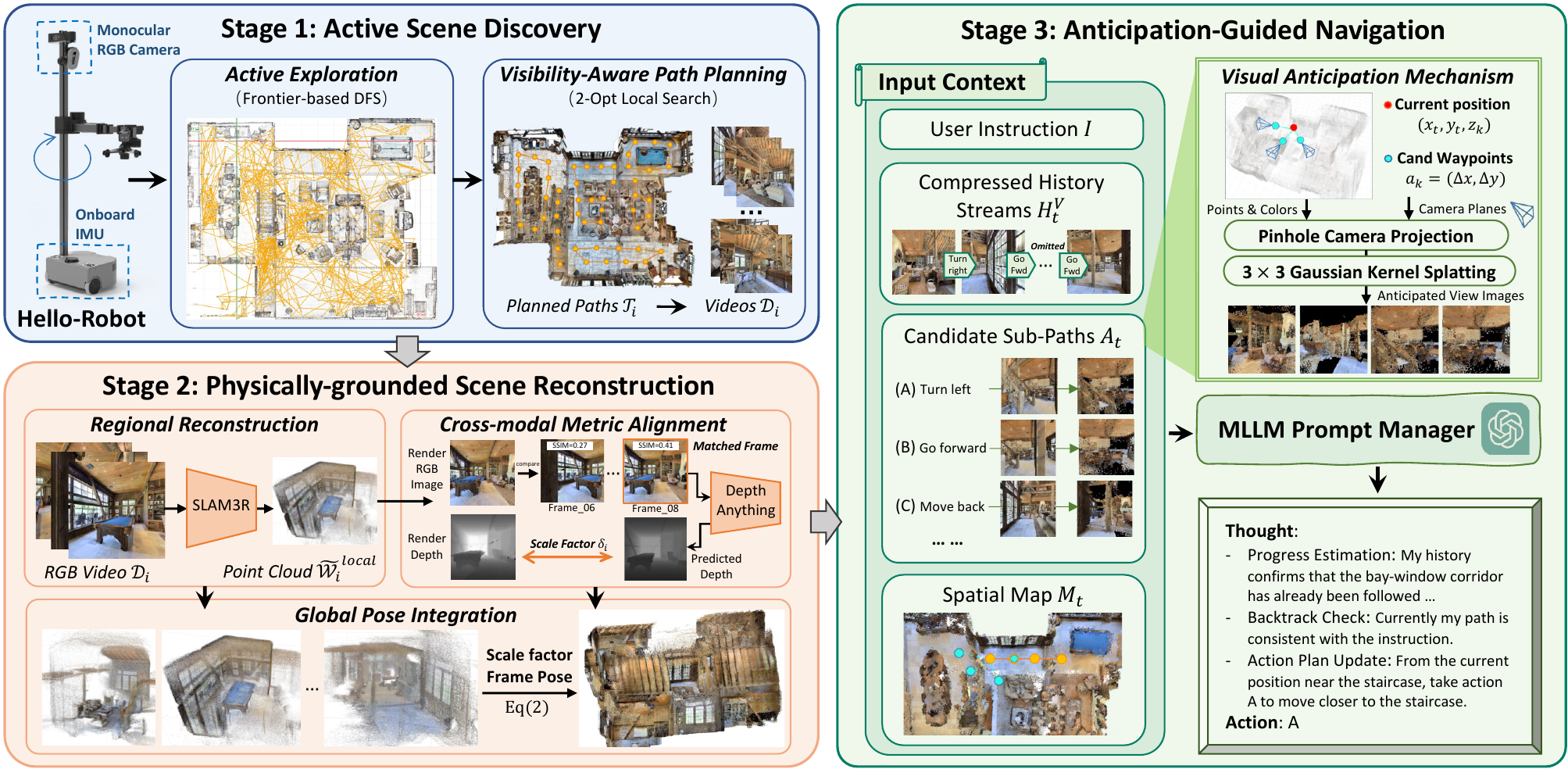}
    \caption{\textbf{Overview of our proposed SpatialAnt framework.} We firstly let the agent to pre-explore the environment, gathering the coordinates of navigable places. Then, we use these information to plan visibility-aware tours to collect geometrically coherent RGB videos and IMU traces for subsequent regional point cloud reconstruction. The reconstructed point clouds are then aligned with physical scale and merged into a global point cloud. Afterall, this global point cloud provides extra global context by a novel visual anticipation mechanism for MLLM to conduct sub-path based high-level reasoning.}
    \label{fig:framework}
\end{figure*}

\section{PROBLEM FORMULATION}
\label{sec:problem_formulation}

In this work, we investigate the task of vision-and-language navigation in continuous environments (VLN-CE), where an embodied agent is required to navigate in a continuous 3D environment $\mathcal{W} \in \mathbb{R}^{3}$, following human instructions $I$ to reach a target destination $(x_d, y_d, z_d)$. 
Following real-world applications such as sweeping robots, we adopt a practical setting where an agent is allowed to pre-explore the environment before task execution, yielding a reconstructed 3D scene point cloud $\widetilde{\mathcal{W}}$. Equipped with this reconstruction, the agent will receive additional signals to serve as global priors for the navigation. Specifically, at time-step $t$ the agent state $\mathcal{S}$ is defined as a quadruple $\textbf{s}_t = (x_t, y_t, z_t, \theta_t)$, where $\theta$ represents the heading of the agent. 
For every step, the agent perceives the environment through panoramic observations by evenly rotating right at a 30 degree interval, yielding 12 RGB images $\mathcal{O}=\{o_1, ..., o_{12}\}$, with each image $o\in \mathbb{R}^{H\times W\times 3}$. The action space $\mathcal{A}$ is defined as a set of continuous local waypoints predicted by a waypoint predictor~\cite{hong2022bridging}, together with an explicit backtrack action as suggested by~\cite{shi2025smartway}. Each action $a_k$ denotes a movement $(\Delta x, \Delta y)$ relative to the agent current position. 
Combining these local perception with additional global signals from the reconstructed scene $\widetilde{\mathcal{W}}$, the zero-shot agents then apply MLLM as the policy model $f$ to predict the next action. 

\section{METHOD}
\label{sec:methods}

We propose \textbf{SpatialAnt}, an autonomous framework that bridges the gap between noisy reconstructions and robust navigation in real scenarios (Figure~\ref{fig:framework}). Specifically, our system has three progressive stages:
(1) \textit{Active scene discovery} that gathers diverse egocentric observations through pre-exploration; 
(2) \textit{Physically-grounded scene reconstruction} that consolidates these observations into a coherent 3D point cloud with real metrics; (3) \textit{Anticipation-guided navigation} that leverages the reconstructed model for rendering future observations to conduct counterfactual reasoning. 


\subsection{\textbf{Active Scene Discovery}}
\label{subsec:scene_discovery}

To autonomously explore the environment and collect high-quality visual data for subsequent scene reconstruction, we introduce an active scene discovery algorithm, as summarized in Algorithm~\ref{alg:discovery}, which is decoupled into two steps: active scene exploration and visibility-aware path planning.

\paragraph{\textbf{Active Scene Exploration}} 
The primary objective of this step is to discover the navigable boundaries of the unseen environment. We initiate the agent from an arbitrary origin $\mathbf{p}_0$.
The agent then performs frontier-based exploration using a depth-first search strategy to incrementally reveal the scene structures. To keep consistent with the navigation, we utilize the waypoint predictor to generate candidate locations for the agent. To manage the continuous 3D space, we discretize all positions into a grid-based navigation graph $\mathcal{G} = (\mathcal{V}, \mathcal{E})$ and apply a non-uniform voxelization to map a position $\mathbf{p}=(x,y,z)$ into a unique grid node id: the horizontal $X-Y$ grid size $\Delta_{xy}$ is set to 0.2m for precise obstacle representation, while the vertical $Z$ grid size $\Delta_{z}$ is set as 0.5m. For each node $v\in \mathcal{V}$, we assign a binary attribute to distinguish between regions the agent has visited and those that are merely observed as candidates. We only add edges between two physically traversable nodes. The exploration process terminates when the set of frontier nodes is exhausted. This ensures that the agent obtains a comprehensive navigation map while minimizing redundant movements. Moreover, by monitoring consecutive vertical movements and applying graph community detection (i.e. Louvain algorithm~\cite{que2015scalable}), we identify stairs and roughly partition the navigation graph into distinct region-level subgraphs $\{\mathcal{G}_1, ...,\mathcal{G}_n\}$.

\begin{algorithm}[t]
\caption{\textbf{Active Scene Discovery}}
\label{alg:discovery}
\begin{algorithmic}[1]
\REQUIRE Origin $\mathbf{p}_0$, Voxel sizes $(\Delta_{xy}, \Delta_{z})$ for Navigation graph $\mathcal{G}$, Coverage threshold $\tau$, Overlap penalty $\lambda_{\text{overlap}}$

\STATE \textbf{Initialize:} $\mathcal{G} \leftarrow (\{\mathbf{p}_0^{\text{visited}=1}\}, \emptyset)$, $\textit{Frontiers} \leftarrow \{\mathbf{p}_0\}$

\vspace{0.3em}
\STATE \textcolor{gray}{\textit{// Step 1: Active Scene Exploration}}
\WHILE{$\textit{Frontiers} \neq \emptyset$}
    \STATE $\mathbf{p}_t \leftarrow \text{DFS}(\textit{Frontiers})$ \textcolor{gray}{\textit{// Move to next place}}
    \STATE $\mathbf{p}_{\text{cand}} \leftarrow \text{WaypointPredictor}(\mathbf{p}_t)$ 
    \STATE $\mathcal{G} \leftarrow \mathcal{G} \,\cup\, (\{\mathbf{p}_{t}^{\text{visited}=1}, \mathbf{p}_{\text{cand}}^{\text{visited}=0}\}, \{\mathbf{p}_{t-1} \leftrightarrow \mathbf{p}_t\})$ 
    \STATE $\textit{Frontiers} \leftarrow \textit{Frontiers} \cup \{v\in \mathcal{G}, v^{\text{visited}=0}\}$
\ENDWHILE
\STATE Identify stairs and partition subgraphs $\{\mathcal{G}_1, \dots, \mathcal{G}_n\}$ 

\vspace{0.3em}
\STATE \textcolor{gray}{\textit{// Step 2: Visibility-Aware Path Planning}}
\FOR{each subgraph $\mathcal{G}_i$}
    \STATE \textcolor{gray}{\textit{// Hub Selection via Weighted Set Cover}}
    \STATE $\mathcal{V}_{hub} \leftarrow \text{VoxelDownsample}(\mathcal{G}_i, \text{grid}=\text{1m})$ 
    \STATE $\mathcal{H} \leftarrow \text{GreedySetCover}(\mathcal{V}_{hub})$ using Eq.(\ref{eq:coverage_threshold}) 
    
    \vspace{0.3em}
    \STATE \textcolor{gray}{\textit{// Path Planning}}
    \STATE $\mathcal{T}_i \leftarrow \text{2-opt}(\mathcal{H})$ using visibility-aware
cost function $C$
    \STATE $\mathcal{T}_i \leftarrow \text{RefineWithSafety}(\mathcal{T}_i, \text{clearance})$ 
\ENDFOR
\\
\STATE \textbf{return} Planned paths $\{\mathcal{T}_1, \dots, \mathcal{T}_n \}$
\end{algorithmic}
\end{algorithm}

\paragraph{\textbf{Visibility-Aware Path Planning}} 
Unlike the stochastic trajectories obtained from active exploration, at this step we aim to collect view-continuous RGB videos with pose traces as the basis for 3D reconstruction, by planning and executing visibility-aware scanning paths that ensure sufficient spatial coverage and view overlap.
For each subgraph $\mathcal{G}_i$, we formulate the path planning task as a weighted visibility-aware traveling salesman problem~\cite{applegate2011traveling}. 
We solve the problem by first downsampling all reachable grid cells by voxel sparsification with grid size set as 1m. 
We then select a subset of nodes as candidate hubs and employ a greedy set cover algorithm to identify a subset of hubs $\mathcal{H}$ that satisfies a volumetric coverage threshold $\tau$:
\begin{equation}
   \frac{|\bigcup_{h\in\mathcal{H}}\text{Vis}(h) \cap \mathcal{G}_i|}{|\mathcal{G}_i|} \ge \tau
\label{eq:coverage_threshold}
\end{equation}
where $\text{Vis}(h)$ is the estimated visible cell set for hub $h$ with an estimated 3m radius. We set $\tau=0.9$ to guarantee sufficient geometric information for high-fidelity reconstruction. Then we plan a path by using a 2-opt local search~\cite{applegate2011traveling} for an optimal Hamiltonian cycle over $\mathcal{H}$ with a visibility-aware cost function $C(u \rightarrow v) = L(u, v) + \lambda_{overlap}(\textit{IoU}(\text{Vis}(u) \rightarrow \text{Vis}(v)))$, 
where $L(u, v)$ is the path distance between two nodes and $\lambda_{overlap}$ is a penalty term for insufficient visual overlap between adjacent hubs.
To ensure safety in real-world deployment and account for the robot's physical body size, the planned paths are refined to maintain a sufficient clearance from obstacles. The agent then traverses each path $\mathcal{T}_i$ with a pre-defined start heading $\theta^{\mathcal{T}_i}_{0}$ using a monocular RGB camera to record a video $\mathcal{D}_i$ together with the agent state traces $\mathcal{S}_i$ from IMU.
At each node on the path, the agent executes a ``look-around'' action by rotating $\pm 60^\circ$ horizontally and $\pm 45^\circ$ vertically, resulting in geometrically consistent RGB videos for the next scene reconstruction.

\subsection{\textbf{Physically-grounded Scene Reconstruction}}
\label{subsec:scene_reconstruction}


To transform the egocentric RGB videos and IMU traces collected during previous active scene discovery into a metric-consistent 3D representation $\widetilde{\mathcal{W}}$ of the real environment $\mathcal{W}$, we design a physically-grounded scene reconstruction pipeline, featuring regional reconstruction, cross-modal metric alignment, and global pose integration. 

\paragraph{\textbf{Regional Reconstruction}} To ensure robustness against lighting variations and potential pose estimation drifts during pre-exploration, we recover the local 3D point cloud $\widetilde{\mathcal{W}}_i^{\textit{local}} \in \mathbb{R}^{K\times 3}$ for each RGB video $\mathcal{D}_i$ by employing SLAM3R \cite{liu2025slam3r}, a state-of-the-art neural SLAM model. For each regional point cloud, the origin is defined at the beginning robot position of the corresponding video. 
Although SLAM3R bypasses the requirements for precise depth maps, the downside of this advantage is that the estimated metric of the reconstructed point cloud is not aligned with the real physical metric. This scale ambiguity, coupled with the lack of a unified global coordinate system, prevents the point clouds from being used directly for navigation.

\paragraph{\textbf{Cross-modal Metric Alignment}} To bridge this gap, we propose a cross-modal alignment approach that anchors the relative depth of $\widetilde{\mathcal{W}}_i^{\textit{local}}$ to an absolute metric scale. Specifically, for each regional point cloud $\widetilde{\mathcal{W}}_i^{\textit{local}}$, we render a forward-facing RGB image $o_f^{i} \in \mathbb{R}^{H\times W \times 3}$ together with a depth map $d_f^{i}\in \mathbb{R}^{H\times W}$ at the local origin. We identify the best-matching video frame $o_t^{i} \in \mathbb{R}^{H\times W \times 3} $ at the start position of path $\mathcal{T}_i$ by computing the average structural similarity score~\cite{wang2004image} across RGB channels. Then, we employ Depth-Anything-2~\cite{yang2024depth} to predict the metric depth $\hat{d}_t\in \mathbb{R}^{H\times W}$ of that frame. The point cloud scale factor $\delta_i$ is then determined by computing the average ratio of the metric depth $\hat{d}_t$ and rendered depth $d_f^{i}\in \mathbb{R}^{H\times W}$ over the set of valid pixels $\Omega$, i.e.  $\delta_i = \sum_{(j,k)\in\Omega} \hat{d}_t(j,k)/d_f^{i}(j,k)$ 
where $\Omega$ excludes pixels from reconstruction holes or depth-missing areas. By applying this scale factor, we obtain the metric-aligned regional point cloud as $\widetilde{\mathcal{W}}_i^{\textit{metric}} = \delta_i \cdot \widetilde{\mathcal{W}}_i^{\textit{local}}$. 

\paragraph{\textbf{Global Pose Integration}}
To construct a unified scene, we further transform the local point clouds into a global coordinate system by utilizing the start position $\mathbf{p}_0^{\mathcal{T}_i}$ and absolute heading $\theta_0^{\mathcal{T}_i}$ of path $\mathcal{T}_i$, together with the relative heading $\theta_t$ corresponding to the matched frame. 
\begin{equation}
\widetilde{\mathcal{W}}_i^{\textit{global}}
=
\left(
\mathbf{R}_z\!\left(\theta_0^{\mathcal{T}_i}+\theta_t\right)
\left(\widetilde{\mathcal{W}}_i^{\textit{metric}}\right)^{\top}
+
\mathbf{p}_0^{\mathcal{T}_i} \mathbf{1}^{\top}
\right)^{\top}
\end{equation}
where $\mathbf{R}_z(\cdot)\in\mathbb{R}^{3\times 3}$ is the rotation matrix relative to the $Z$-axis, and $\mathbf{1}\in\mathbb{R}^{3}$ is an all-ones vector.
Finally, we have the reconstructed point cloud for the whole scene by merging all the local point clouds together, $\widetilde{\mathcal{W}} = \bigcup_{i=1}^{N} \widetilde{\mathcal{W}}_i^{\textit{global}}$. 

\subsection{\textbf{Anticipation-Guided Navigation}}
\label{subsec:antipication_nav}

To enable MLLMs to reason with explicit scene knowledge, we propose a novel navigation paradigm that efficiently transforms the point clouds into MLLM-compatible formats. Specifically, given the following components: (1) the Instruction $I$, (2) the agent current position $\mathbf{p}_t=(x_t, y_t, z_t)$ and heading $\theta_t$ in point cloud $\widetilde{\mathcal{W}}$ coordinate system, (3) the visual Observations $\mathcal{O}_t$, (4) the Action Space $\mathcal{A}_t$, (5) the History $H_t$, which is a succinct record of the agent’s previous actions. 
We define an anticipation-guided navigation paradigm that endows the MLLM agent with the ability of ``foreseeing before arriving'' by introducing a visual anticipation mechanism for counterfactual reasoning via future sub-path selection. 

\paragraph{\textbf{Visual Anticipation Mechanism}} 
Unlike generative world models that \textit{\textbf{imagine}} future frames through stochastic synthesis~\cite{wang2025imaginenav++}, which suffer from physical hallucinations, our approach \textit{\textbf{anticipates}} future perspectives via a deterministic geometric projection from our physically-grounded scene point clouds. Specifically, for each navigable waypoint $a_k=(\Delta x_k, \Delta y_k) \in \mathcal{A}_t$, we calculate its absolute position in the point cloud coordinate system $\mathbf{p}_{a_k}=(x_t + \Delta x_k, y_t + \Delta y_k, z_k)$. To calculate the expected visual observation $\hat{o}_k \in \mathbb{R}^{256\times 256\times 3}$ at this navigable location with anticipated heading $\phi_k=\text{atan2}(\Delta y_k, \Delta x_k)$, we adopt a gaussian splatting algorithm~\cite{yang2024spec}. For each point $\mathbf{p} \in \widetilde{\mathcal{W}}$, we firstly project it onto the target image plane using the pinhole camera model. 
Then to mitigate the possible noise and structural holes, we apply a $3\times 3$ gaussian kernel to distribute the color and depth of each point to its neighboring pixels.
To handle occlusions, only the points whose depth falls within a small tolerance $\epsilon$ of the minimum recorded depth at a given pixel contribute to the final color through a distance-weighted accumulation. 

\paragraph{\textbf{Counterfactual Reasoning via Sub-Path Selection}} Instead of most previous VLN agents that take the navigation as the next waypoint select problem~\cite{qiao2025open,shi2025smartway,li2026one}, we formulate the navigation as a sub-path selection problem. Specifically, at each step we provide the MLLM with:
\begin{itemize}
    \item Compressed History Streams ($H_t^V$): To maintain the path coherence, we represent the historical trajectory as a temporal stream of visual observations, connected by the actions taken. As the raw history sequences have much redundancy, particularly related to long-range forward movements, we therefore introduce a simple compression strategy by merging the intermediate steps for consecutive forward actions. We only retain the key observations that mark the beginning and end of the forward sequence. The remaining history streams are further resized to $256 \times 256$ pixel size. This strategy allows the MLLM to perceive the history as a series of coherent events without overwhelming the context.
    \item Anticipated Navigable Sub-Paths ($A_k$): We augment each candidate action $a_k$ with a projected future view $\hat{o}_k$ by using our visual anticipation mechanism. By presenting these anticipated views alongside the current observation, we transform the action decision into a counterfactual visual alignment task by asking the MLLM to select the sub-path (together with the history) that best aligns with the instruction. Specifically, each candidate action has the form: ``[observed view $o_k$] [action $a_k$] $\rightarrow$ if I took this action, I would partially imagine the next view: [anticipated view $\hat{o}_k$]''. This form allows the MLLM to perform semantic cross-verification on the physical reality of potential future observation, hence reducing the error backtracking.
    \item Spatial Map ($M_t$): Following~\cite{zhang2026spatialnav}, we additionally provide a top-down topological representation to ground the agent location in a global spatial context. The map is generated by projecting the reconstructed 3D point cloud $\widetilde{\mathcal{W}}$ onto a 2D occupancy grid. To provide the MLLM with a coherent understanding of the navigation progress, we also visualize the historical trajectory and the candidate waypoints together on the map.
\end{itemize}
Combining these components with the user instruction $I$, the MLLM (specifically, GPT-5.1-2025-11-13) acts as a high-level spatial-temporal policy that evaluates the semantic consistency of each sub-path candidate $A_k$:
\begin{equation}
    A^* = f \left( 
    I, H_t^V, \mathcal{O}_k, \{A_k|a_k\in \mathcal{A}_t\}, M_t 
    \right)
\end{equation}
After selecting the best $A^*$, the agent executes the corresponding action $a_k$ and moves to the next place.

\begin{table*}[t]
\centering
\small
\captionsetup{skip=2pt}
\setlength{\tabcolsep}{5pt}
\setlength{\abovecaptionskip}{2pt}
\setlength{\belowcaptionskip}{3pt}

\caption{\textbf{Comparison in continuous environments on R2R-CE and RxR-CE Val-Unseen splits.} 
The \textbf{best supervised} results are highlighted in \textbf{bold}, while the {best zero-shot} results are \underline{underlined}. ``Pre-Exp'' denotes whether the zero-shot agent adopts the pre-exploration based navigation settings.}
\label{tab:continuous_main}

\resizebox{0.98\linewidth}{!}{
\begin{tabular}{clc ccccc cccc}
\toprule

\multirow{2}{*}{\textbf{\#}} & 
\multirow{2}{*}{\textbf{Methods}} &
\multirow{2}{*}{\textbf{Pre-Exp}} &
\multicolumn{5}{c}{\textbf{R2R-CE}} & \multicolumn{4}{c}{\textbf{RxR-CE}} \\
\cmidrule(lr){4-8} \cmidrule(lr){9-12} 
& & & \textbf{NE}($\downarrow$) & \textbf{OSR}($\uparrow$) & \textbf{SR}($\uparrow$) & \textbf{SPL}($\uparrow$) & \textbf{nDTW}($\uparrow$) & \textbf{NE}($\downarrow$) & \textbf{SR}($\uparrow$) & \textbf{SPL}($\uparrow$) & \textbf{nDTW}($\uparrow$) \\
\midrule
\rowcolor{blue!5} \multicolumn{12}{l}{\textbf{\textit{Supervised Learning:}}} \\

1 & NavFoM~\cite{zhang2025embodied} & -- & 4.61 & 72.1 & 61.7 & 55.3 & -- & 4.74 & 64.4 & \textbf{56.2} & 65.8 \\
2 & Efficient-VLN~\cite{zheng2025efficient} & -- & \textbf{4.18} & \textbf{73.7} & \textbf{64.2} & \textbf{55.9} & -- & \textbf{3.88} & \textbf{67.0} & 54.3 & \textbf{68.4} \\
\midrule
\rowcolor{blue!5} \multicolumn{12}{l}{\textbf{\textit{Zero-Shot:}}} \\
3 & Open-Nav~\cite{qiao2025open} & $\times$ & 6.70 & 23.0 & 19.0 & 16.1 & 45.8 & -- & -- & -- & -- \\
4 & Smartway~\cite{shi2025smartway} & $\times$ & 7.01 & 51.0 & 29.0 & 22.5 & -- & -- & -- & -- & -- \\
5 & STRIDER~\cite{he2025strider} & $\times$ & 6.91 & 39.0 & 35.0 & 30.3 & 51.8 & 11.19 & 21.2 & 9.6 & 30.1 \\
6 & VLN-Zero~\cite{bhatt2025vln}& $\checkmark$  & 5.97 & 51.6 & 42.4 & 26.3 & -- & 9.13 & 30.8 & 19.0 & -- \\
7 & SpatialNav~\cite{zhang2026spatialnav} & $\checkmark$ &
5.15 & 66.0 & 64.0 & 51.1 & 65.4
& 7.64 & 32.4 &  24.6 & 55.0 \\

\midrule 
\rowcolor{gray!10} 
8 & \textbf{SpatialAnt (Ours)} & $\checkmark$  & 
\underline{4.42} & \underline{76.0} & \underline{66.0} & \underline{54.4} & \underline{69.5} &
\underline{5.28} & \underline{50.8} & \underline{35.6} & \underline{65.4} \\

\bottomrule
\end{tabular}
}

\end{table*}

\begin{table}[t]
\vspace{-15pt}

\centering
\small
\captionsetup{skip=2pt}
\setlength{\tabcolsep}{5pt}
\setlength{\abovecaptionskip}{2pt}
\setlength{\belowcaptionskip}{3pt}

    \caption{\textbf{Comparison of backtrack times on the successful tasks.} We report the results by agent-reconstructed scene point clouds on sampled val-unseen splits of R2R-CE. }
    \label{tab:analy_backtrack}

    \resizebox{0.95\linewidth}{!}{
    \begin{tabular}{l|c|ccc}
    \toprule
    \textbf{Methods}    &       \textbf{\#Backtrack}  & \textbf{\#Steps(↓)} & \textbf{TL(↓)} & \textbf{SPL(↑)} \\
    \midrule
    SpatialNav~\cite{zhang2026spatialnav}  & 8 / 51 (15.7\%) & 8.60 & 10.77 & 77.9 \\
    \midrule
    \textbf{SpatialAnt (Ours)} & 8 / 66 (\textbf{12.1\%}) & \textbf{8.26} & \textbf{10.48} & \textbf{84.9} \\
    \bottomrule
    \end{tabular}
    }
    
\hfill
\vspace{-5pt}
\end{table}

\section{EXPERIMENTS}

\subsection{\textbf{Experiment Setup}}

\paragraph{\textbf{Simulation Benchmarks}}
We evaluate our method in the Habitat simulator~\cite{puig2023habitat} using the R2R-CE~\cite{krantz2020beyond} and RxR-CE datasets~\cite{ku2020room}. 
These datasets provide complex and diverse instructions in the continuous environment spanning 11 distinct scans. To align with established evaluation protocols~\cite{qiao2025open,shi2025smartway,zhang2026spatialnav}, we randomly sample 100 and 195 episodes from the validation unseen split of each dataset respectively for computationally intensive zero-shot MLLM agents. To keep consistent with previous works, we adopt the simulator-rendered metric depth for waypoint predictor~\cite{shi2025smartway}.

\paragraph{\textbf{Real-World Deployment}}
For real-world validation, we deploy our model on a Hello Robot platform~\cite{HelloRobot2025}. The robot is equipped with an Intel RealSense D435if RGB camera for visual perception and a 9-DoF IMU. 
We adopt Depth-Anything-2\cite{yang2024depth} to produce the metric depth used by waypoint predictor. All onboard inference is executed on a single NVIDIA RTX A6000 GPU. 
For fair comparison, we follow the experimental protocols defined in \cite{shi2025smartway,shi2025fast} to construct a real-world dataset consisting of 25 cross-room paths featuring open-vocabulary targets, such as a ``red ladder''. As introduced in Section~\ref{subsec:scene_discovery} and Section~\ref{subsec:scene_reconstruction}, we leverage the robot's onboard sensors to pre-explore and reconstruct the environment before navigation. This setup evaluates the transition of our approach from high-fidelity simulation to unstructured real-world scenarios.

\paragraph{\textbf{Evaluation Metrics}}
Following the standard protocols for VLN-CE~\cite{hong2022bridging}, we evaluate our model using the following metrics: {Success Rate (SR)}, {Success weighted by Path Length (SPL)}, {Navigation Error (NE)}, {Normalized Dynamic Time Warping (nDTW)}, and {Trajectory Length (TL)}. 
Following previous approaches in \cite{qiao2025open,shi2025smartway,he2025strider}, an episode is considered successful if the robot stops within a distance threshold $d_{th}$ from the goal, where $d_{th} = 3.0$m in simulation and $d_{th} = 2.0$m for real-world experiments to reflect the higher precision required in physical environments.

\paragraph{\textbf{Baselines}}
We evaluate SpatialAnt against two categories of competitive methods: (1) \textit{Supervised Learning} approaches, including current state-of-the-art {NavFoM}~\cite{zhang2025embodied} and {Efficient-VLN}~\cite{zheng2025efficient}, and (2) \textit{Zero-Shot} methods, spanning exploration-free navigators (including {Open-Nav}~\cite{qiao2025open}, {Smartway}~\cite{shi2025smartway} and {STRIDER}~\cite{he2025strider}) and pre-exploration based approaches (including {VLN-Zero}~\cite{bhatt2025vln} and {SpatialNav}~\cite{zhang2026spatialnav}). Our SpatialAnt is directly comparable with the latter two methods under a shared pre-exploration setting.

\subsection{\textbf{Main Results}}
Table~\ref{tab:continuous_main} provides a quantitative comparison of our proposed SpatialAnt with state-of-the-art (SOTA) methods in R2R-CE and RxR-CE datasets. The results demonstrate a clear trend that pre-exploration based agents significantly outperform normal on-the-fly zero-shot agents, highlighting the necessity of explicitly modeling the global spatial knowledge.

On R2R-CE, SpatialAnt reaches the supervised-level performance with a 66\% SR, slightly exceeding the strongest supervised baselines Efficient-VLN~\cite{zheng2025efficient}. This indicates that the presence of global scene priors reduces the dependence on in-domain training to achieve SOTA performance. Moreover, SpatialAnt improves significantly over the previous SOTA zero-shot baseline, SpatialNav~\cite{zhang2026spatialnav}, by raising the OSR from 66\% to 76\% while increasing the SPL and nDTW to 54.4\% and 69.5\% respectively. This efficiency gain is further supported by Table~\ref{tab:analy_backtrack}, SpatialAnt executes 3.6\% less backtracking than SpatialNav to succeed, indicating fewer erroneous moves and more faithful instruction following.

On the more challenging RxR-CE, where previous methods suffer severe performance degradation, SpatialAnt maintains excellent generalization with 50.8\% SR and 35.6\% SPL, significantly surpassing prior zero-shot methods by +18.4\% SR and +11.0\% SPL. Such improvement indicates that our anticipation-based sub-path selection provides a robust reasoning basis for MLLM against the complex cross-room instructions, thereby enabling high path efficiency.

\begin{table}[t]
\vspace{-15pt}

\centering
\small
\captionsetup{skip=2pt}
\setlength{\tabcolsep}{3pt}
\setlength{\abovecaptionskip}{2pt}
\setlength{\belowcaptionskip}{3pt}
\caption{\textbf{Comparison between the human-curated (H) and agent-reconstructed (A) scene point clouds.}}

\label{tab:ablation_pcd}
\centering
\resizebox{\linewidth}{!}{
\begin{tabular}{c|c|ccc|ccc}
\toprule
\multirow{2}{*}{\textbf{Methods}} & 
\multirow{2}{*}{\textbf{Scene}} & 
\multicolumn{3}{c|}{\textbf{R2R-CE}} & 
\multicolumn{3}{c}{\textbf{RxR-CE}} \\
\cmidrule{3-8}
&  & \textbf{OSR(↑)} & \textbf{SR(↑)} & \textbf{SPL(↑)} & \textbf{OSR(↑)} & \textbf{SR(↑)} & \textbf{SPL(↑)} \\
\midrule
\multirow{2}{*}{SpatialNav~\cite{zhang2026spatialnav}} & 
  H & 66.0 & 64.0 & 51.1 & 47.5 & 32.4 & 24.1 \\
& A & 61.0 & 51.0 & 39.8 & 40.3 & 26.1 & 19.3 \\
\midrule
\rowcolor{blue!5} 
\multicolumn{2}{l|}{\textbf{\textit{Performance gap}}} & 
-5.0 & -13.0 & -11.3 & -7.2 & -6.3 & -4.8
\\
\midrule
\multirow{2}{*}{\textbf{SpatialAnt}} 
 & H & 80.0 & 69.0 & 52.7 & 69.2 & 53.9 & 37.6 \\
 & A & 76.0 & 66.0 & 54.4 & 63.1 & 50.8 & 35.6 \\
\midrule
\rowcolor{blue!5} 
\multicolumn{2}{l|}{\textbf{\textit{Performance gap}}} & 
-4.0 & -3.0 & -1.7 & -6.1 & -2.1 & -2.0
\\
\bottomrule
\end{tabular}
}
\hfill
\vspace{-5pt}
\end{table}

\subsection{\textbf{Ablation Study}}
We conduct comprehensive ablation studies to validate the robustness and effectiveness of our framework.

\paragraph{\textbf{Robustness to Sim-to-Real Transfer}} To investigate how reconstruction noise affects exploration-based zero-shot agents, we compare SpatialNav~\cite{zhang2026spatialnav} and our SpatialAnt using:
\begin{itemize}
    \item Human-curated (H) high-quality point clouds provided by Matterport3D dataset~\cite{chang2017matterport3d}, which is an idealized pre-exploration assumption taken by SpatialNav that perfect scene point clouds are directly available,
    \item Agent-constructed (A) point clouds using our methods.
\end{itemize}
As presented in Table~\ref{tab:ablation_pcd}, the performance of SpatialNav drops significantly (-13.0\% SR on R2R-CE and -6.3\%SR on RxR-CE) when switching to noisy reconstructed point clouds, indicating that using spatial-map only as an absolute reference yields poor robustness to real scenarios. In contrast, SpatialAnt exhibits much smaller degradation with a -3.0 \% SR on R2R-CE and a -2.1\% on RxR-CE), because our SpatialAnt treats the reconstructed geometry as a source of counterfactual visual evidence by constructing verifiable sub-paths, thereby effectively narrowing the sim-to-real gap.

\begin{table}[t]
\centering
\small
\captionsetup{skip=2pt}
\setlength{\tabcolsep}{3pt}
\setlength{\abovecaptionskip}{2pt}
\setlength{\belowcaptionskip}{3pt}
\caption{\textbf{Ablation study on visual anticipation mechanism (VA) and its horizon field of view (HFOV).} We report the results on sampled val-unseen splits of R2R-CE.}
\label{tab:ablation_ant}
\centering
\resizebox{\linewidth}{!}{
\begin{tabular}{l|c|ccccc}
\toprule
\textbf{Methods} & \textbf{HFOV} & \multicolumn{1}{l}{\textbf{NE(↓)}} & \textbf{OSR(↑)} & \textbf{SR(↑)} & \textbf{SPL(↑)} & \multicolumn{1}{l}{\textbf{nDTW(↑)}} \\
\midrule
SpatialAnt (w/o VA) & - & 5.19 & 77.0 & 55.0 & 41.4 & 70.0 \\
\midrule
\multirow{3}{*}{SpatialAnt (w/ VA)} 
 & 60$^\circ$ & \textbf{4.33} & 72.0 & 65.0 & \textbf{55.2} & 69.2 \\
 & 90$^\circ$ & 4.42 & \textbf{76.0} & \textbf{66.0} & 54.4 & \textbf{69.5} \\
 & 120$^\circ$ & 4.85 & 74.0 & 59.0 & 48.7 & 67.7 \\
\bottomrule
\end{tabular}
}
\hfill
\end{table}

\begin{figure}[t]
\setlength{\abovecaptionskip}{-1pt}
\setlength{\belowcaptionskip}{0pt}
\centering
    \includegraphics[width=\linewidth]{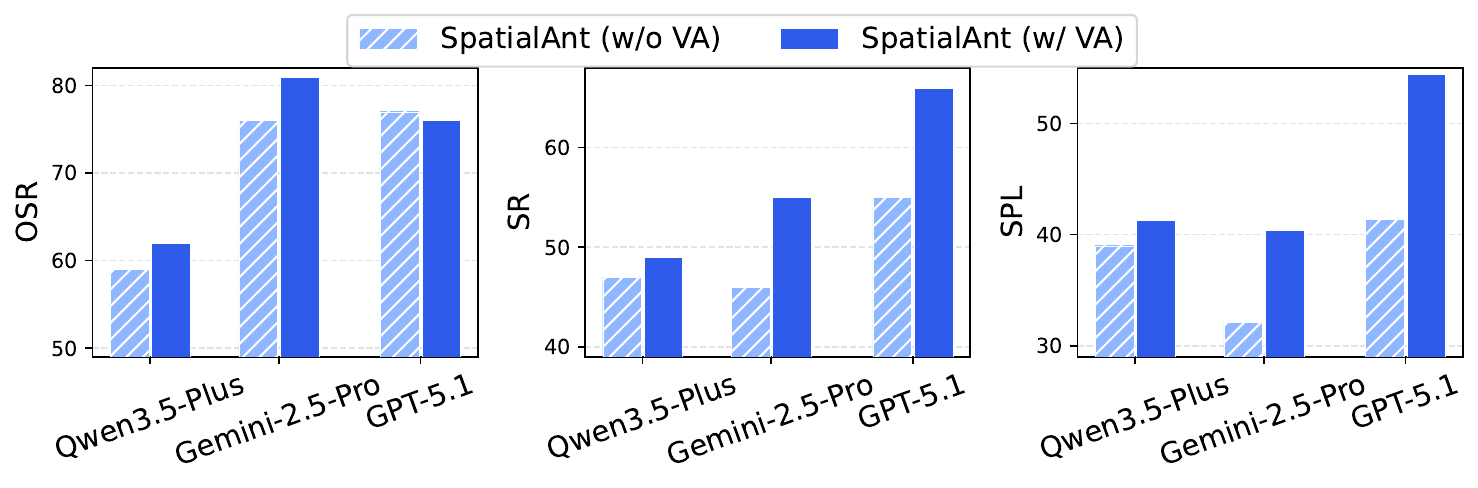}
    \caption{\textbf{Comparison across different MLLM backbones.} We report the results on sampled subset of R2R-CE. The default HFOV of visual anticipation (VA) is set as 90$^\circ$.}
    \label{fig:llm_ablation}
\end{figure}

\begin{table}[t]
    \centering
    \small
    \captionsetup{skip=2pt}
    \setlength{\tabcolsep}{3.5pt}
    \setlength{\abovecaptionskip}{2pt}
    \setlength{\belowcaptionskip}{3pt}
    
    \caption{\textbf{Real-world navigation performance.} We compare SpatialAnt against representative supervised-learning (VLN-BERT) and zero-shot (SmartWay) baselines. 
    }
    \label{tab:real_world_results}
    \resizebox{\linewidth}{!}{
    \begin{tabular}{l cccc}
        \toprule
        \textbf{Method} & \textbf{LP Time}(s) & \textbf{GP Time}(s) & \textbf{SR}$(\uparrow$, \%) & \textbf{NE} ($\downarrow$, m) \\
        \midrule
        
        \rowcolor{blue!5} \multicolumn{5}{l}{\textbf{\textit{Supervised Learning:}}} \\
        VLN-BERT~\cite{hong2021vln} & 22.40 & - & 20.0 & 3.92 \\        
        \midrule
        
        \rowcolor{blue!5} \multicolumn{5}{l}{\textbf{\textit{Zero-Shot:}}} \\
        SmartWay~\cite{shi2025smartway} & 22.40 & - & 32.0 & 3.01 \\ 
        
        \textbf{SpatialAnt (Ours)} & 22.40 & 0.57 & \textbf{52.0} & \textbf{2.49} \\
        \bottomrule
    \end{tabular}
    }
\end{table}

\begin{figure}[t]
\setlength{\abovecaptionskip}{0pt}
\setlength{\belowcaptionskip}{0pt}
\centering  

    \begin{subfigure}{\linewidth}
        \centering
        \includegraphics[width=\linewidth]{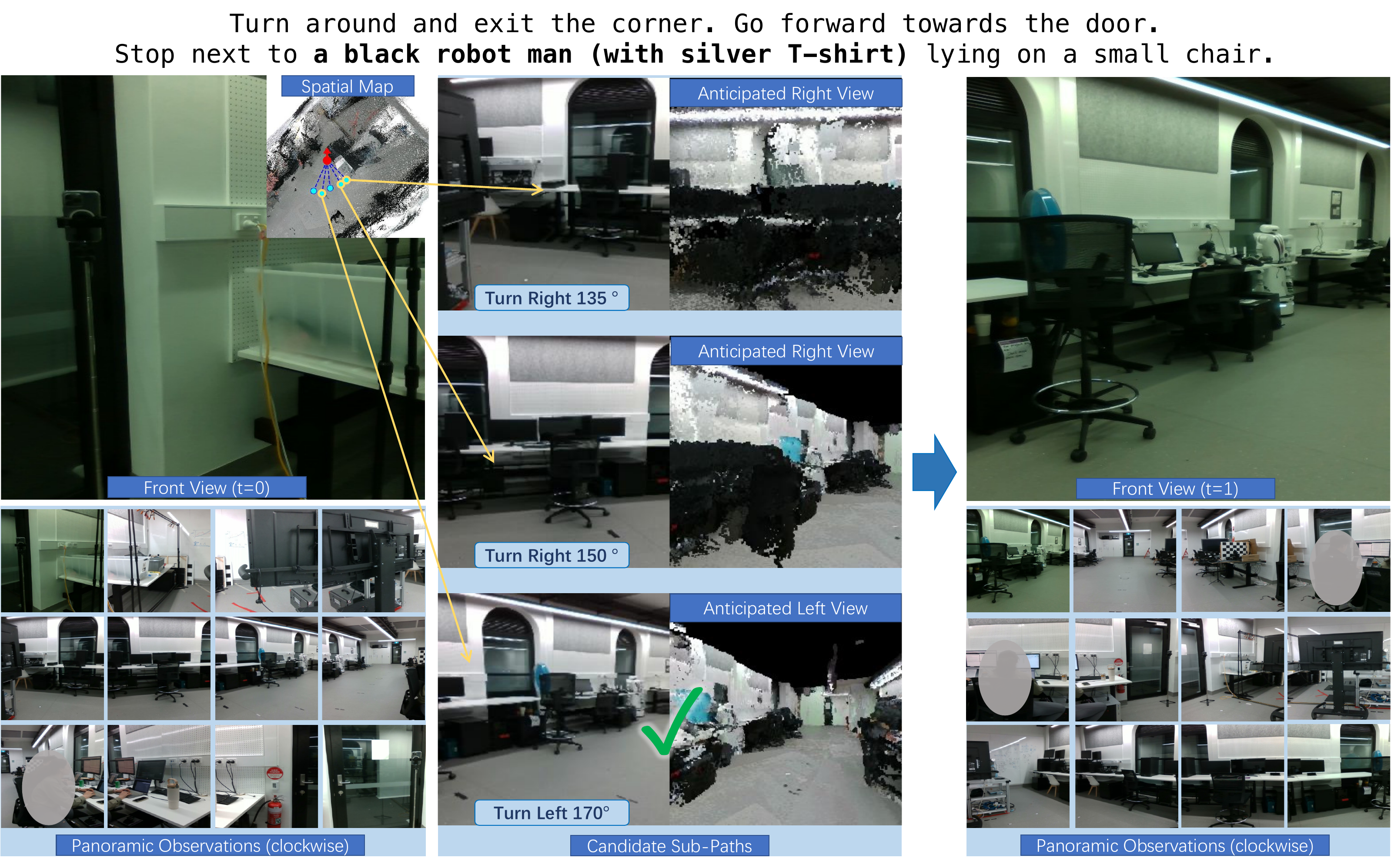}
        \vspace{-0.6cm}
        \caption{Example: Disambiguate at the beginning.}
        \label{fig:case_study_a}
    \end{subfigure}

    \vspace{0.1cm} 

    \begin{subfigure}{\linewidth}
        \centering
        \includegraphics[width=\linewidth]{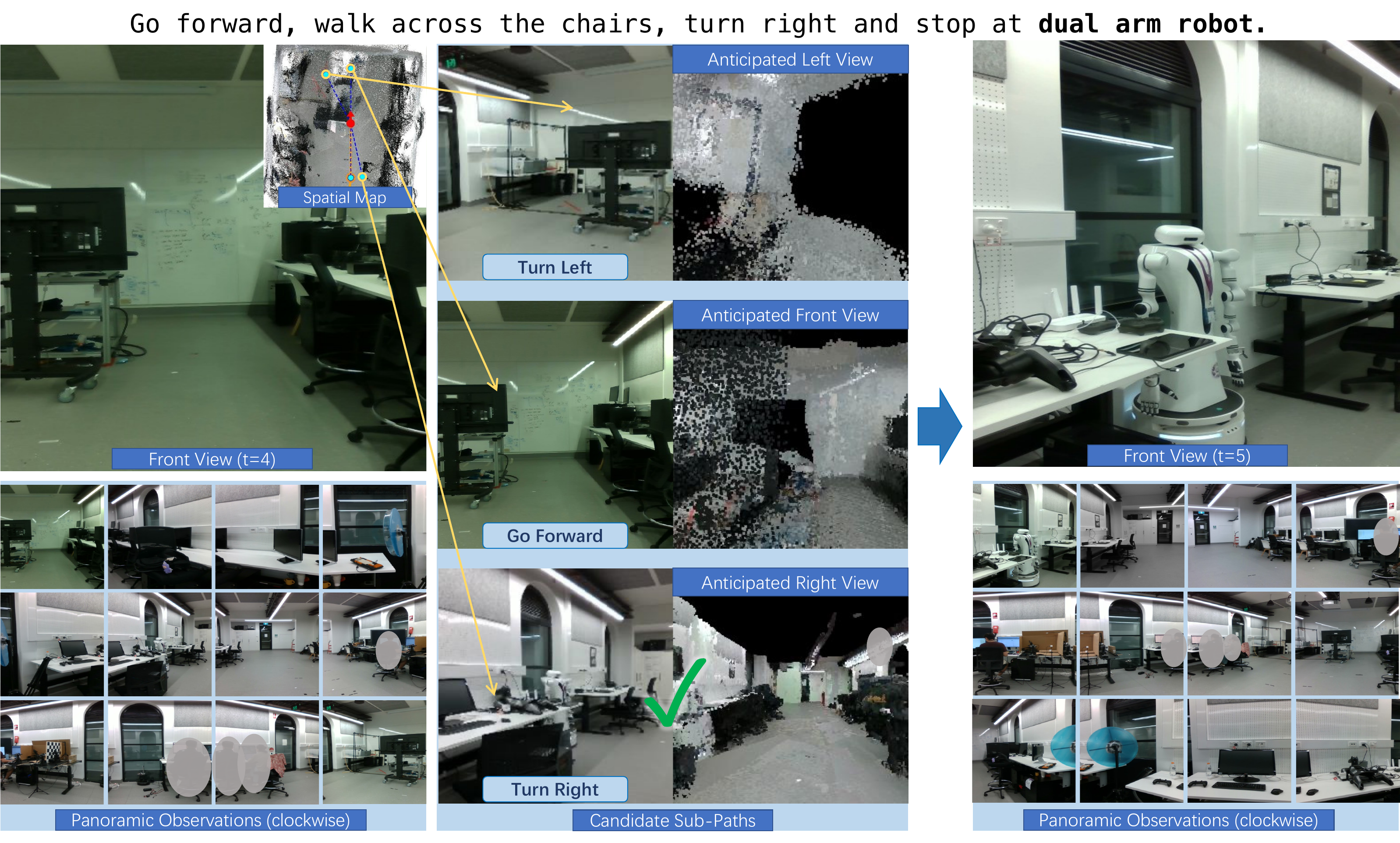}
        \vspace{-0.6cm}
        \caption{Example: Amend at midway.}
        \label{fig:case_study_b}
    \end{subfigure}

    \vspace{0.0cm}
    \caption{\textbf{Representative cases in real environments.} We show frames of the agent egocentric front view, panoramic observations (from right to left), and candidate actions on the spatial map together with our anticipated views. }
    \label{fig:case_study}
\end{figure}

\paragraph{\textbf{Effectiveness of Visual Anticipation Mechanism}}
To verify the necessity and impact of visual anticipation (VA), we compare the performance of our agent with and without this mechanism. Table~\ref{tab:ablation_ant} demonstrates that incorporating visual anticipation consistently improves most evaluation metrics, confirming that VA provides crucial guidance for action disambiguation. We further ablate the anticipation horizon field of view (HFOV). Results illustrate that HFOV set as 90$^\circ$ achieves the best overall trade-off, offering richer look-ahead evidence that 60$^\circ$ HFOV, while avoiding the increased uncertainty introduced by a wider 120$^\circ$ HFOV. 

\paragraph{\textbf{Scalability with MLLM Reasoning Abilities}} We validate the model-agnostic property of our framework by instantiating SpatialAnt with multiple MLLM backbones. As shown in Figure~\ref{fig:llm_ablation}, we can have two key insights. Firstly, the effectiveness of our visual anticipation mechanism is model-agnostic. Secondly, the improvements on SpatialAnt become more pronounced as the reasoning capability of the backbone increases, showing that stronger MLLMs are better at leveraging the counterfactual evidence provided by anticipated observations to disambiguate actions and avoid instruction-inconsistent paths. These results suggest that SpatialAnt scales favorably with future MLLM advances, translating improved reasoning capacity into navigation gains.

\subsection{\textbf{Real-world Deployment}}
To validate the practical effectiveness of our framework, we deploy SpatialAnt in the real environment with Hello-Robot and conduct pre-exploration. The video collection following the planned path finished within 10 minutes per region, resulting in around 1000 frames. The scene reconstruction then took about 200 seconds on a single NVIDIA RTX A6000 GPU. 
Besides the navigation metrics, we also report the per-step local perception (LP) time that used to collect real visual observations, and the per-step global perception (GP) time that used to project spatial map and gather visual anticipations from the point cloud. As shown in Table~\ref{tab:real_world_results}, our method achieves significantly higher success rate along with the reduced navigation error, indicating that our pre-explored, physically-grounded scene priors provide actionable guidance in real scenarios. Importantly, with the reconstructed global point cloud (which is a one-time cost), the global perception is very efficient, demonstrating that the additional computation introduced by our framework is lightweight and well within practical deployment constraints.

\subsection{\textbf{Case Study}}

We present two qualitative examples in Figure~\ref{fig:case_study} to demonstrate how SpatialAnt benefits from the global priors obtained by pre-exploration. As shown in Figure~\ref{fig:case_study_a}, at the beginning, the candidate waypoints all seem plausible for the ``turn around'' instruction. In this case, SpatialAnt queried the pre-explored point cloud to render anticipated observations, resulting in semantically different sub-paths, and selected the one that best supports the subsequent instruction ``go forward towards the door''. The example in Figure~\ref{fig:case_study_b} further shows SpatialAnt’s ability to recover from an early mistake: after the agent passed the dual-arm robot (because the waypoint predictor fails to propose a proper candidate), our rendered sub-paths provide counterfactual evidence that highlights the dual-arm robot and triggers corrective action to re-align with the instruction.
Taken together, these cases illustrate that the practical value of our framework extends beyond mere increased performance, but to equip the agent with complex reasoning skills in anticipated physical space.


\section{CONCLUSION}
In this work, we present \textbf{SpatialAnt}, a zero-shot VLN framework that closed-loop the scene exploration to scene reconstruction, finalizing the real robotic navigation. By actively exploring the environment with a single RGB camera, SpatialAnt introduces a physical grounding strategy that resolves monocular scale ambiguity to produce metric-consistent point clouds. Rather than treating the self-constructed point clouds only as reference maps, we propose a visual anticipation mechanism that renders future observations from this noisy geometry and supports counterfactual reasoning to validate candidate sub-paths against instructions. Extensive results on R2R-CE and RxR-CE, together with real-world deployment on the Hello Robot, demonstrate superior zero-shot performance under imperfect scene reconstructions, highlighting that reliable navigation depends on how the world is grounded and verified.

\bibliographystyle{icml2026}
\bibliography{papers}

\end{document}